\documentclass[conference]{IEEEtran}

\IEEEoverridecommandlockouts

\usepackage{graphicx}  
\usepackage{cite}  
\usepackage{amsmath,amssymb,amsfonts}  
\usepackage{algorithmic}  
\usepackage{textcomp}  
\usepackage{xcolor}  
\usepackage{array}  
\usepackage{graphicx}  
\usepackage{lipsum}  
\usepackage{subcaption}
\usepackage[utf8]{inputenc}  
\usepackage{url}  
\usepackage{hyperref}  
\usepackage{booktabs}  
\usepackage{cite} 
\usepackage{array} 

\begin{document}

\title{Predictive Analytics for Dementia: Machine Learning on Healthcare Data}
\author{
    \IEEEauthorblockN{1\textsuperscript{st} Shafiul Ajam Opee}
    \IEEEauthorblockA{\textit{Faculty of Science \& Technology} \\
    \textit{American International University-Bangladesh} \\
    Dhaka, Bangladesh \\
    opee.cse@gmail.com}
    \and
    \IEEEauthorblockN{2\textsuperscript{nd} Nafiz Fahad}
    \IEEEauthorblockA{\textit{Faculty of Information Science \& Technology} \\
    \textit{Multimedia University} \\
    Melaka, Malaysia \\
    fahadnafiz1@gmail.com}
    \and
    \IEEEauthorblockN{3\textsuperscript{rd} Anik Sen}
    \IEEEauthorblockA{\textit{Faculty of Information Science \& Technology} \\
    \textit{Multimedia University} \\
    Melaka, Malaysia \\
    aniksen360@gmail.com}
    \and 
    \IEEEauthorblockN{4\textsuperscript{th} Rasel Ahmed}
    \IEEEauthorblockA{\textit{Faculty of Computing \& Informatics} \\
    \textit{Multimedia University} \\
    Cyberjaya, Malaysia \\
    raselahmed1337@gmail.com}
    \and
    \IEEEauthorblockN{5\textsuperscript{th} Fariha Jahan}
    \IEEEauthorblockA{\textit{Faculty of Science \& Technology} \\
    \textit{American International University-Bangladesh} \\
    Dhaka, Bangladesh \\
    fariha.rainy23@gmail.com}
    
    \and
    \IEEEauthorblockN{6\textsuperscript{th} Md. Kishor Morol}
    \IEEEauthorblockA{\textit{Faculty of Science \& Technology} \\
    \textit{American International University-Bangladesh} \\
    Dhaka, Bangladesh \\
    kishoremorol@gmail.com}
    \and \\

    \IEEEauthorblockN{7\textsuperscript{th}Md Rashedul Islam}
    \IEEEauthorblockA{\textit{Department of CSE} \\
    \textit{University of Asia Pacific, Bangladesh,} \\
    Dhaka, Bangladesh \\
    rashed.cse@upa-bd.edu}
}

\maketitle
\begin{abstract}
Dementia is a complex syndrome impacting cognitive and emotional functions, with Alzheimer's disease being the most common form. This study focuses on enhancing dementia prediction using machine learning (ML) techniques on patient health data. Supervised learning algorithms are applied in this study, including K-Nearest Neighbors (KNN), Quadratic Discriminant Analysis (QDA), Linear Discriminant Analysis (LDA), and Gaussian Process Classifiers. To address class imbalance and improve model performance, techniques such as Synthetic Minority Over-sampling Technique (SMOTE) and Term Frequency-Inverse Document Frequency (TF-IDF) vectorization were employed. Among the models, LDA achieved the highest testing accuracy of 98\%. This study highlights the importance of model interpretability and the correlation of dementia with features such as the presence of the APOE-$\epsilon$4 allele and chronic conditions like diabetes. This research advocates for future ML innovations, particularly in integrating explainable AI approaches, to further improve predictive capabilities in dementia care.
\end{abstract}


\section{INTRODUCTION}
Dementia significantly affects cognitive and emotional functions, impairing memory, decision-making abilities, communication, and attention span \cite{ref1, ref2}. It is an umbrella term that encompasses various cognitive disorders, with Alzheimer's disease being the most common type \cite{ref3}. Dementia manifests differently in each individual, and its likelihood increases with age, particularly affecting those over the age of 65 \cite{ref4}. In 2018, it was estimated that 50 million people were living with dementia globally, a number that has now exceeded 55 million, with over 60\% residing in low-income countries. This growing burden imposes substantial financial costs, estimated at over 1 trillion USD annually, and these expenses are expected to rise as the population ages \cite{ref5}.

The causes of dementia are multifactorial, influenced by social, economic, and geographic factors \cite{ref6}. Links between dementia and conditions such as depression, diabetes, smoking, physical inactivity, midlife obesity, and hyperlipidemia have been well established \cite{ref8}. Moreover, dementia risk is classified into low, moderate, and high-risk categories, based on factors such as advanced age, lower educational levels, and lower income. Addressing modifiable risk factors—such as engaging in physical activity and improving cardiovascular health—can help reduce the likelihood of developing dementia, as noted by the World Dementia Council (WDC) \cite{ref11}. Currently, dementia ranks as the seventh leading cause of death and is a significant cause of disability and dependency among older adults globally \cite{ref10}.

In response to the rising prevalence and associated challenges of dementia, innovative approaches are required, particularly in diagnostics. Traditional methods, while valuable, are often insufficient for early detection. Machine learning (ML) has emerged as a promising solution, offering the potential to predict dementia more accurately by leveraging large datasets. ML approaches are divided into two categories: supervised and unsupervised learning.  Supervised learning creates models from labelled data, whereas unsupervised learning clusters or separates data based on inherent patterns. \cite{ref11}.

The rest of this article is organized into several key sections. The Literature Review discusses the application of ML in predicting cognitive diseases, highlighting relevant studies and methodologies. The Methodology section details data collection, preprocessing, feature engineering, and model training. The Proposed Architecture explains the ML classifiers and hybrid neural network models. Data Preprocessing covers techniques for handling missing values and encoding categorical variables. The ML Models section discusses classifiers like KNN, LDA, QDA, and Gaussian Process Classifiers. The Deep Learning and Hybrid Approaches section introduces a hybrid model combining CNN and RNN. Finally, the Results section presents the study’s findings, and the Discussion analyzes their clinical significance.

\section{LITERATURE REVIEW}
ML has become a powerful tool in the field of cognitive disease prediction, often surpassing traditional statistical methods due to its ability to handle large datasets and complex patterns. Here, in this section, existing works in the field of supervised learning, especially deep learnign and the key insights from existing works have been discussed. 
\subsection{Supervised Learning Techniques}
Herzog and Magoulas \cite{ref11} explored the use of supervised learning models, particularly Convolutional Neural Networks (CNN), for distinguishing between Normal Cognition (NC) and Early Mild Cognitive Impairment (EMCI). They achieved impressive accuracies of 92.5\% and 75.0\%, respectively, for distinguishing NC from EMCI and 93.0\% and 90.5\% for distinguishing NC from Alzheimer's Disease (AD). Their findings highlighted the ability of CNNs to capture complex patterns in cognitive data. Irfan et al. \cite{ref12} combined cognitive data and neuroimaging features in an AdaBoost Ensemble model, achieving an accuracy of 83\% in early-stage dementia detection. Their study emphasized the need for new ML methodologies to further enhance detection efficiency and suggested that integrating multiple data sources could improve prediction accuracy. Rajayyan and Mustafa \cite{ref13} utilized Gaussian Naive Bayes coupled with a cuckoo algorithm for performance improvement. Their model reached 95\% accuracy, with precision and recall scores of 97\% and 95\%, respectively, demonstrating the importance of feature engineering and the benefits of optimization algorithms in dementia prediction. Similarly, Vyshnavi et al. \cite{ref14} applied artificial neural networks (ANNs) and random forest techniques on a large NHATS dataset containing over 5000 individuals aged over 60. Initially, the model achieved an accuracy of 68\%, but after using random forest classification on balanced and raw data, the accuracy improved to 92\%. This work underscores the value of refining data preprocessing techniques to improve model performance. Rawat et al. \cite{ref20} proposed a stacked model using Gradient Boosting Machine and ANN. Their model attained 89\% accuracy, illustrating that combining models can enhance prediction systems by leveraging the strengths of individual algorithms. Dhakal et al. \cite{ref16} applied several supervised classifiers, including KNN, Support Vector Machines (SVM), and Decision Trees (DT), on the OASIS dataset. The SVM classifier outperformed others, achieving 96.77\% accuracy, highlighting the importance of feature selection and preprocessing in achieving high-performance metrics. Their future plans involve implementing ensemble techniques and feature reduction to further improve model reliability.

\subsection{Deep Learning and Hybrid Approaches}
Deep learning has shown significant potential in dementia prediction. Bidani et al. \cite{ref17} developed an innovative deep learning model that integrates Deep Convolutional Neural Networks (DCNNs) with transfer learning. The DCNN model alone achieved an accuracy of 81.94\%, while the transfer learning method resulted in a lower accuracy of 68.13\%. This suggests that although deep learning is effective for feature extraction, its effectiveness is strongly influenced by the type of data used and the architecture of the model. Javeed et al. \cite{ref21} carried out an in-depth review of ML-based automated diagnostic systems that utilize a range of data types, including images, clinical features, and voice data. Their analysis provided valuable insights into the various ways ML can be applied across different data types for dementia prediction, emphasizing the importance of customizing models to fit the specific characteristics of the dataset.

\subsection{Key Insights from Existing Work}
The studies reviewed demonstrate that supervised learning, particularly with feature engineering and optimization techniques, can achieve high accuracy in dementia prediction. Methods such as ensemble learning and model stacking have been shown to improve model performance. However, the integration of multiple data types (e.g., cognitive, neuroimaging, and clinical data) remains a challenge and an area of ongoing research. Moreover, while deep learning approaches show promise, their success is highly dependent on the data used, and further work is needed to refine these models for clinical applications.

\begin{table*}[!ht]
\centering
\caption{Comparison of ML Techniques for Dementia Prediction}
\label{tab:ml_techniques}
\begin{tabular}{>{\raggedright}m{2.5cm}>{\raggedright}m{2.5cm}>{\raggedright\arraybackslash}m{2.5cm}>{\raggedright\arraybackslash}m{2.5cm}>{\raggedright\arraybackslash}m{5.5cm}}
\toprule
\textbf{Study} & \textbf{Technique} & \textbf{Data Type} & \textbf{Accuracy (\%)} & \textbf{Key Insights} \\ 
\midrule
Herzog \& Magoulas \cite{ref11} & CNNs & Cognitive Data & 92.5 - 93.0 & High accuracy in distinguishing cognitive states \\ 
Irfan et al. \cite{ref12} & AdaBoost Ensemble & Cognitive \& Neuroimaging & 83.0 & Early-stage detection, combined data sources \\ 
Rajayyan \& Mustafa \cite{ref13} & Gaussian Naive Bayes + Cuckoo Algorithm & Cognitive Data & 95.0 & High precision and recall, feature engineering \\ 
Zadgaonkar et al. \cite{ref14} & ANN \& Random Forest & NHATS Dataset & 68.0 - 92.0 & Improved accuracy with random forest, large dataset \\ 
Rawat et al. \cite{ref20} & Stacked Gradient Boosting \& ANN & Various & 89.0 & Enhanced accuracy through model stacking \\ 
Dhakal et al. \cite{ref16} & KNN, SVM, DT & OASIS Dataset & 96.77 & Highest accuracy with SVM, feature selection importance \\ 
Bidani et al. \cite{ref17} & DCNN + Transfer Learning & MRI Images & 68.13 - 81.94 & Deep learning for feature extraction \\ 
Javeed et al. \cite{ref21} & Various ML Techniques & Images, Clinical, Voice & N/A & Comprehensive assessment across data types \\ 
\bottomrule
\end{tabular}
\end{table*}

\section{METHODOLOGY}
Our methodology involves a detailed and structured approach to predict dementia using ML techniques. This process encompasses several key stages, including data collection, preprocessing, feature engineering, model training, and evaluation.

\subsection{Data Collection}
The "Dementia Patient Health and Prescriptions Dataset," available on Kaggle, is an extensive dataset developed for ML and data science tasks focused on dementia prediction and analysis. It includes a wide range of features related to the health and prescription information of dementia patients, such as demographic details (age, gender, educational background, and family history of dementia), health metrics (e.g., diabetic status, alcohol consumption, heart rate, blood oxygen level, body temperature, weight, and MRI delay), and prescription data (e.g., medication names and dosages). The dataset also covers lifestyle factors (e.g., smoking habits, physical activity levels, diet, and sleep quality) and genetic data (e.g., presence of the APOE-$\epsilon$4 allele). Additionally, it contains cognitive test scores, depression status, medication history, and details about chronic health conditions (such as diabetes, heart disease, and hypertension). Despite the dataset's comprehensive nature, the Kaggle website does not provide descriptions of these features, which means users must explore the dataset themselves to fully understand each variable. To address this limitation, the current study employed the "Dementia Patient Health and Prescriptions Dataset" from Kaggle, which includes features like demographic information (age, gender, education), health parameters (e.g., diabetes status, heart rate, blood oxygen level), and genetic information (e.g., presence of the APOE-$\epsilon$4 allele). We performed exploratory data analysis to ensure a thorough understanding of the dataset's variables \cite{ref19}.

\subsection{Proposed Architecture}
\begin{figure}[h!]
\raggedright
\hspace{1cm}
\includegraphics[width=0.46\textwidth]{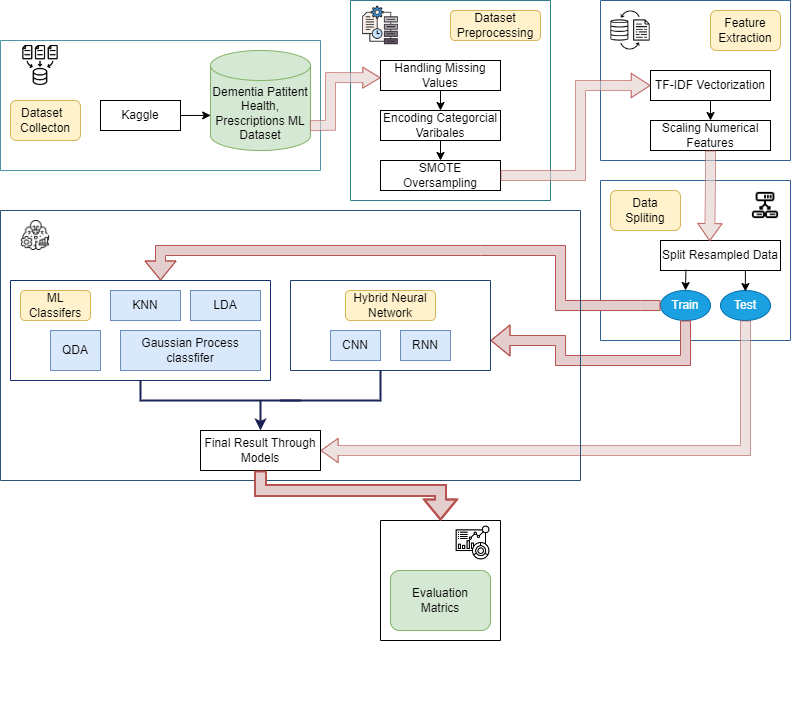}
\caption{Proposed Architecture}
\label{fig:architecture}
\end{figure}

Figure \ref{fig:architecture} shows that in the initial phase, this current study meticulously collected and prepared our dataset using several data preprocessing and essential feature engineering techniques. Following standardization, we implemented various supervised ML classifiers, including QDA, KNN, Naive Bayes (NB), Ridge, and Gaussian Process Classifiers, to predict dementia. This comprehensive approach ensures that the data is well-prepared and that the models are robust and accurate. The diagram in Figure 1 provides a visual representation of our proposed methodology.

\subsection{Data Preprocessing}
Data preprocessing plays a crucial role in the ML pipeline, significantly influencing the performance and accuracy of predictive models. This phase involves converting raw data into a clean, well-organized format that is suitable for analysis. In our study, we applied various data preprocessing techniques to guarantee the quality and integrity of the dataset, as outlined below:

\subsubsection{Handling Missing Values}
Missing values in a dataset can negatively impact ML models if not addressed appropriately. In our dataset, some columns contained missing values. To handle this, we employed the \texttt{SimpleImputer} class from the \texttt{sklearn} library, using the 'most\_frequent' strategy. This approach fills in missing values with the most frequently occurring value within each column, maintaining the consistency and integrity of the data.

\subsubsection{Encoding Categorical Variables}
Transforming data into a format that ML models can process is an essential step in ML. Significant categorical variables were converted into non-numeric values using the \texttt{LabelEncoder()} function \cite{ref18}. Additionally, the \texttt{OneHotEncoder()} technique was applied to convert categorical data into numerical form, enabling its integration into ML models.

\subsubsection{Smote Oversampling}
To address class imbalance, we applied Synthetic Minority Over-sampling (SMOTE). This technique generates synthetic samples for the minority class, offering a better representation of the dataset. By using the \texttt{SMOTE()} function and specifying a \texttt{random\_state}, we ensured consistency across reproducible iterations. The process involved resampling both the feature matrix and the target variable, leading to a more balanced distribution across all classes.

\subsection{Feature Engineering}

\subsubsection{TF-IDF Vectorization}
Term Frequency-Inverse Document Frequency (TF-IDF) vectorization was employed using the \texttt{TfidfVectorizer()} function. By iterating over every text column, we applied TF-IDF vectorization to both the training and testing datasets. The vectorizer was first fitted to the training data to learn the vocabulary, and then it was used to transform both the training and testing data accordingly. This process helped in converting textual data into numerical form, making it suitable for ML models.

\subsubsection{Scaling Numerical Features}
To ensure data standardization and compatibility, numerical features were scaled using the \texttt{StandardScaler()} function. This scaling technique standardizes features by removing the mean and scaling them to unit variance, which helps in improving the performance and convergence speed of many ML algorithms.

\subsection{Overview of Algorithms}

\subsubsection{KNN}
The k Nearest Neighbor (KNN) method is commonly recognized for its efficiency and effectiveness in tasks like classification and clustering over various kinds of datasets \cite{ref4}. KNN delivers predictions depending on the average value or majority class of the nearest neighbors by finding the k nearest data points in the feature space to the input query point. 

KNN has distinct distance metrics to work with continuous and categorical variables. With the most popular Euclidean distance metrics \eqref{eq1}, there are two more distances which are Manhattan \eqref{eq2} and Minkowski \eqref{eq3}, commonly used for continuous variables. Hamming distance \eqref{eq4} quantifies the similarity and dissimilarity between two categorical data points by tallying the attributes \cite{ref6}.

\paragraph{Euclidean Distance:}
\begin{equation}
d(x, y) = \sqrt{\sum_{i=1}^{n} (x_i - y_i)^2} \label{eq1}
\end{equation}

\paragraph{Manhattan Distance:}
\begin{equation}
d(x, y) = \sum_{i=1}^{n} |x_i - y_i| \label{eq2}
\end{equation}

\paragraph{Minkowski Distance:}
\begin{equation}
d(x, y) = \left( \sum_{i=1}^{n} |x_i - y_i|^p \right)^{1/p} \label{eq3}
\end{equation}

\paragraph{Hamming Distance:}
\begin{equation}
d(x, y) = \sum_{i=1}^{n} \text{diff}(x_i, y_i) \label{eq4}
\end{equation}
Where $n$ is the length of the string and $\text{diff}(x_i, y_i)$ is a function that equals 0 if $x_i = y_i$, and 1 otherwise.

\subsubsection{Linear Discriminant Analysis}
LDA is widely recognized and utilized for classification and dimensionality reduction tasks across various domains \cite{ref7}. To calculate a discriminant profile, the LDA classification score $L_{ij}$ for a specific class $k$ is determined using the following equation, under the assumption that the covariance matrices across classes are identical:
\begin{equation}
L_{ik} = (x_i - \bar{x}_k)^T \Sigma_{\text{pooled}}^{-1} (x_i - \bar{x}_k) - 2 \log_e(\pi_k) \label{eq5}
\end{equation}
Where $x_i$ denotes the measurement vector for sample $i$, $\bar{x}_k$ represents the mean measurement vector of class $k$, $\Sigma_{\text{pooled}}$ is the pooled covariance matrix, and $\pi_k$ is the prior probability of class $k$.

\subsubsection{Quadratic Discriminant Analysis}
In QDA, the classification score $Q_{ij}$ for each class $k$ is computed using the following formula, which involves the variance-covariance matrix for that class:
\begin{equation}
Q_{ij} = (x_i - \bar{x}_k)^T \Sigma_k^{-1} (x_i - \bar{x}_k) + \log_e(|\Sigma_k|) - 2 \log_e(\pi_k) \label{eq6}
\end{equation}
Here, $\Sigma_k$ is the variance-covariance matrix for class $k$, and $\log_e(|\Sigma_k|)$ represents the natural logarithm of the determinant of $\Sigma_k$. The prior probability, pooled covariance matrix $\Sigma_{\text{pooled}}$, and the variance-covariance matrix $\Sigma_k$ are calculated as follows:
\begin{equation}
\pi_k = \frac{N_k}{N} \label{eq7}
\end{equation}
\begin{equation}
\Sigma_{\text{pooled}} = \frac{1}{N} \sum_{k=1}^{K} N_k \Sigma_k \label{eq8}
\end{equation}
\begin{equation}
\Sigma_k = \frac{1}{N_k} \sum_{i=1}^{N_k} (x_i - \bar{x}_k) (x_i - \bar{x}_k)^T \label{eq9}
\end{equation}
Where $N_k$ is the number of samples in class $k$, $N$ is the total number of samples in the training set, and $k$ represents the total number of classes \cite{ref8,ref9}.

\subsubsection{Gaussian Process (GP)}
Gaussian Process (GP) models are effective tools for Bayesian classification \cite{ref10}. In the context of binary classification, we describe the Gaussian process model for classifying a label $y \in \{-1, 1\}$ based on an input $x$. The model is discriminative in that it estimates $p(y|x)$, which, for a fixed $x$, follows a Bernoulli distribution. The probability of a positive outcome, $p(y=1|x)$, is related to an unobserved latent function $f(x)$, which is mapped to the unit interval using a sigmoid function such as the logit or the probit. To simplify calculations, especially for the Expectation Propagation (EP) algorithm, the probit model is utilized:
\begin{equation}
p(y=1|x) = \Phi(f(x))
\end{equation}
Where $\Phi$ represents the cumulative distribution function of the standard normal distribution\cite{ref11}.

\subsection{Hybrid Neural Network}
\begin{figure}[h!]
\raggedright
\hspace{-1.5cm}
\includegraphics[width=0.56\textwidth]{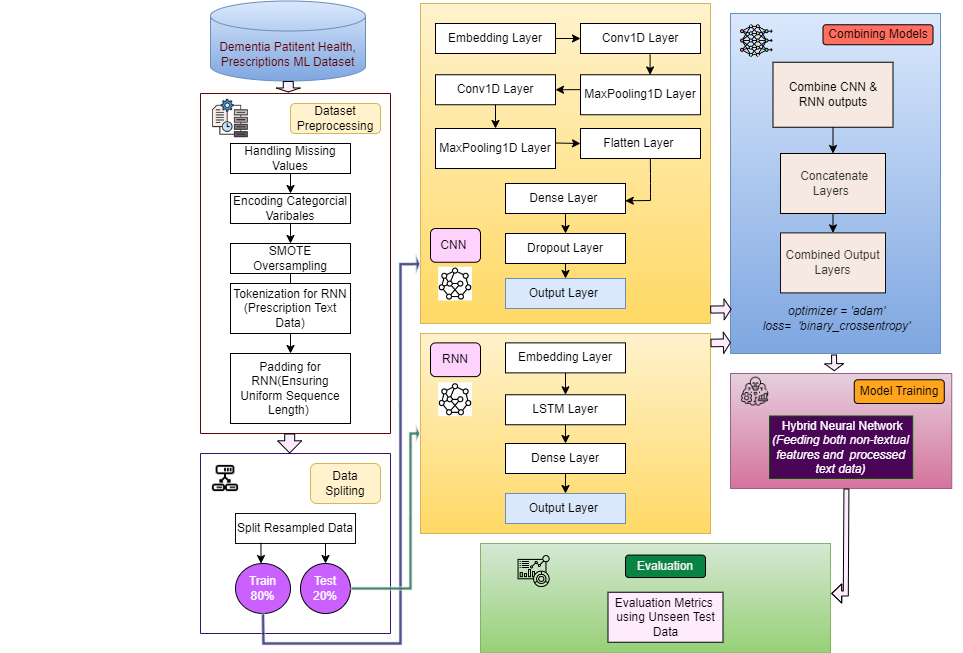}
\caption{Methodological Description of Hybrid Neural Network}
\label{fig:hybrid_network}
\end{figure}

In this approach, we design a hybrid neural network model that integrates the strengths of CNN and Recurrent Neural Networks (RNN) to predict dementia using various patient health data, family history, prescription text data, and other features. The CNN architecture is used to handle non-textual numerical data. The model starts with an embedding layer that transforms input tokens into dense vectors of fixed size. This is followed by two \texttt{Conv1D} layers that perform convolution operations to identify local patterns in the data. These layers are then followed by \texttt{MaxPooling1D} layers, which reduce the dimensionality of each spatial element, aiding in abstracting the presence of features. The output is then passed through a dense layer with a \texttt{ReLU} activation function, introducing non-linearity, and is flattened into a one-dimensional vector. A dropout layer is added to mitigate overfitting by randomly setting a portion of the input units to zero at each update during the training process. The final layer of the CNN model uses a \texttt{sigmoid} activation function, providing a probability value for binary classification tasks.

Prescription text data are processed by the RNN architecture. The text is tokenized at the start, converting it into sequences of integers. These sequences are then padded to ensure uniform input length. The RNN model begins with an embedding layer that converts the tokens into dense vectorized outputs. The processed data is then passed through an \texttt{LSTM} layer, which helps learn long-term dependencies in sequential data. The output of the LSTM layer is passed through another dense layer with a \texttt{sigmoid} activation function, resulting in the RNN’s final prediction. The outputs from both the CNN and RNN models are concatenated to form the hybrid model. This concatenated input is passed through a dense layer to merge the features learned from both models. The final output layer again uses a \texttt{sigmoid} activation function to determine the final prediction. This hybrid model is built using the Adam optimizer and the binary cross-entropy loss function, making it ideal for binary classification tasks. During training, the complete model is fed both the non-textual features and the processed text data, and the loss function is optimized to improve model performance.

\section{RESULTS}

This study explores ML models for dementia prediction using patient health data. This current study focuses on features such as the APOE-$\epsilon$4 allele and chronic conditions like diabetes. Preprocessing methods like SMOTE and TF-IDF were applied, revealing a strong association between genetic and chronic disease factors in dementia risk prediction.

\begin{figure}[h]
    \centering
    \includegraphics[width=0.5\textwidth]{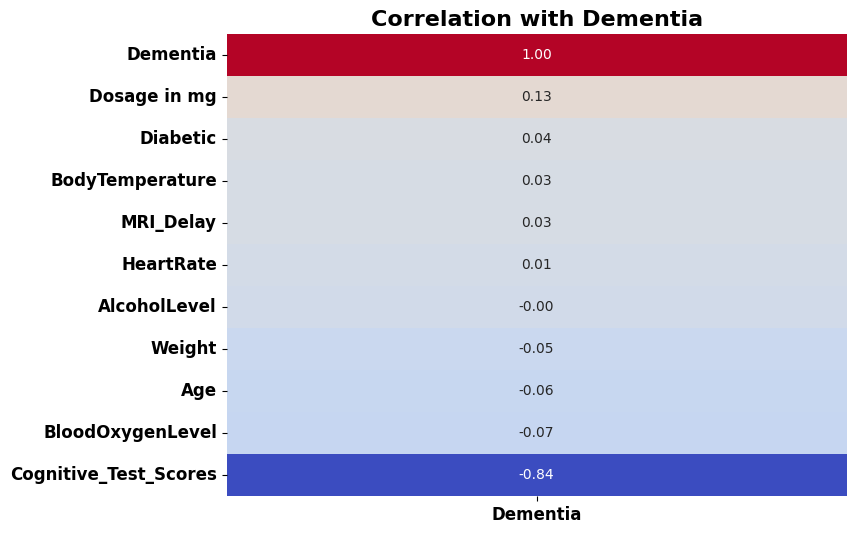}
    \caption{Correlation between Dementia and Features.}
    \label{fig:correlation}
\end{figure}

Figure \ref{fig:correlation} presents the correlation matrix showing the strength and direction of relationships between dementia and various features in the dataset. Correlation values range between -1 and +1, where positive values represent a direct relationship and negative values represent an inverse relationship. In this analysis, the highest positive correlation (0.13) is observed between dementia and the medication dosage, suggesting a potential association between dosage and the onset of dementia. On the other hand, dementia exhibits a strong negative correlation (-0.84) with cognitive test scores, indicating that lower scores are strongly associated with dementia. These insights help in identifying critical predictive features in the model \cite{ref15}.

\begin{figure}[h]
    \centering
    \includegraphics[width=0.3\textwidth]{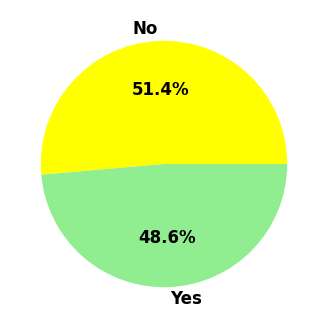}
    \caption{Family history of dementia in our dataset.}
    \label{fig:family_history}
\end{figure}

Figure \ref{fig:family_history} visualizes the distribution of dementia occurrence among patients based on their family history. According to the dataset, 51.4\% of the patients do not have a family history of dementia, while 48.6\% do. This finding implies that family history alone is not a definitive predictor of dementia risk, as the numbers are almost evenly split between those with and without a family history of the disease. The graph demonstrates the need for a more comprehensive analysis of other risk factors when assessing dementia likelihood.

\begin{figure}[h!]
    \centering
    \includegraphics[width=0.44\textwidth]{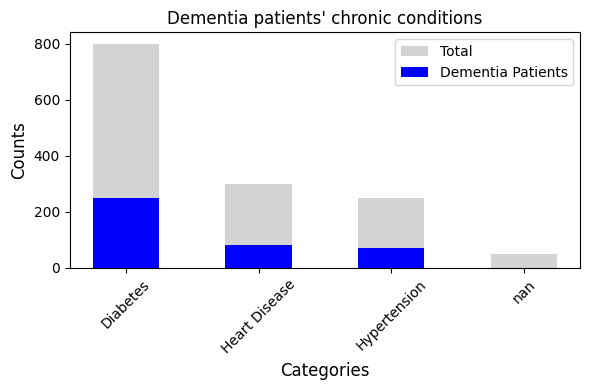}
    \caption{Dementia Patients' Chronic Conditions.}
    \label{fig:chronic_conditions}
\end{figure}
Figure \ref{fig:chronic_conditions} provides a breakdown of chronic conditions among dementia patients. The figure clearly shows that diabetes is the most common comorbidity in patients with dementia, with around 260 out of 760 diabetes patients also diagnosed with dementia. Heart disease and hypertension are also prevalent but are associated with fewer dementia cases compared to diabetes. This indicates that managing diabetes and similar metabolic conditions could play a vital role in dementia prevention or early intervention strategies.

\begin{figure}[h!]
    \centering
    \includegraphics[width=0.44\textwidth]{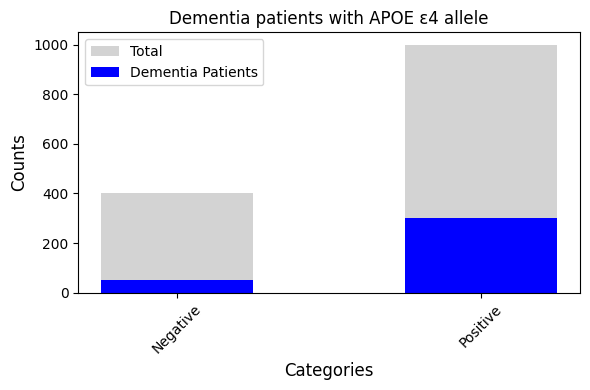}
    \caption{Dementia Patients with APOE-$\epsilon$4 allele.}
    \label{fig:apoe_e4}
\end{figure}

Figure \ref{fig:apoe_e4} highlights the impact of the APOE-$\epsilon$4 allele, a significant genetic risk factor for Alzheimer’s disease and dementia. The figure shows that patients carrying this allele are much more likely to develop dementia than those without it. Approximately 1000 patients in the dataset are APOE-$\epsilon$4 positive, with about 420 of them diagnosed with dementia, while a smaller fraction of patients who do not carry the allele develop the disease. This figure emphasizes the importance of genetic screening in dementia risk assessment, as those with APOE-$\epsilon$4 are at substantially higher risk.

\begin{figure*}[ht] % Use figure* for double-column figures
    \centering
    
    % First Row
    \begin{subfigure}{0.40\textwidth}
        \centering
        \includegraphics[width=\linewidth]{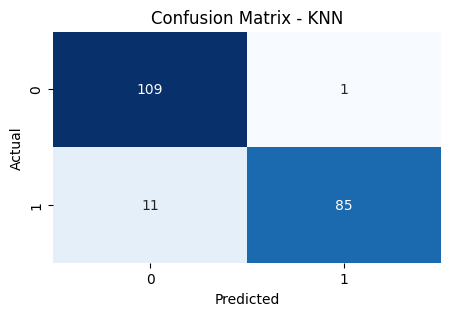}
        \caption{}
        \label{fig:subfig_a}
    \end{subfigure}
    \hspace{0.05\textwidth} % Adjust space between figures
    \begin{subfigure}{0.40\textwidth}
        \centering
        \includegraphics[width=\linewidth]{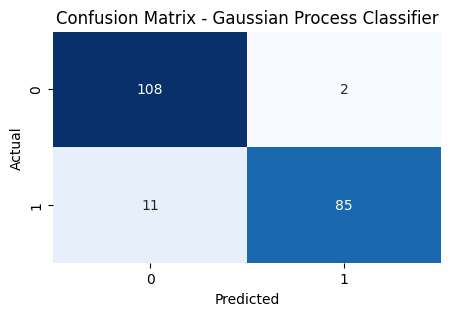}
        \caption{}
        \label{fig:subfig_b}
    \end{subfigure}

    \vspace{0.1cm} % Adjust vertical space between rows
    
    % Second Row
    \begin{subfigure}{0.40\textwidth}
        \centering
        \includegraphics[width=\linewidth]{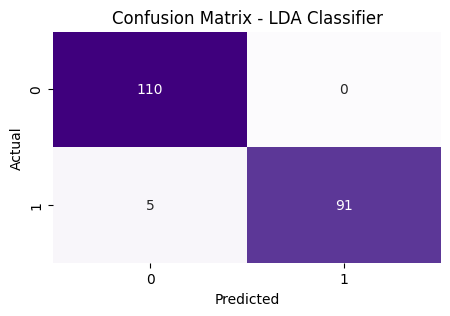}
        \caption{}
        \label{fig:subfig_c}
    \end{subfigure}
    \hspace{0.05\textwidth} % Adjust space between figures
    \begin{subfigure}{0.40\textwidth}
        \centering
        \includegraphics[width=\linewidth]{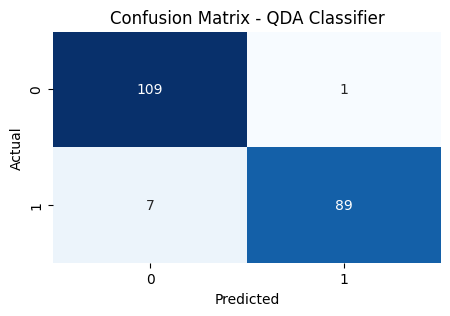}
        \caption{}
        \label{fig:subfig_d}
    \end{subfigure}

    \caption{Confusion Matrix of Four ML Classifiers.}
    \label{fig:confusion_matrices}
\end{figure*}

Figure \ref{fig:confusion_matrices} provides the confusion matrices for KNN, QDA, LDA, and GPC classifiers. Each matrix shows the number of true positives (TP), true negatives (TN), false positives (FP), and false negatives (FN). These metrics are critical for evaluating the classifiers' performance. Among the models, LDA stands out with the highest number of correctly predicted cases (TP and TN) and the lowest number of false negatives, demonstrating its superiority in predicting dementia accurately. KNN, QDA, and GPC also perform well, but their results are slightly lower in terms of precision and recall.

\begin{figure}[h!]
    \centering
    \includegraphics[width=0.5\textwidth]{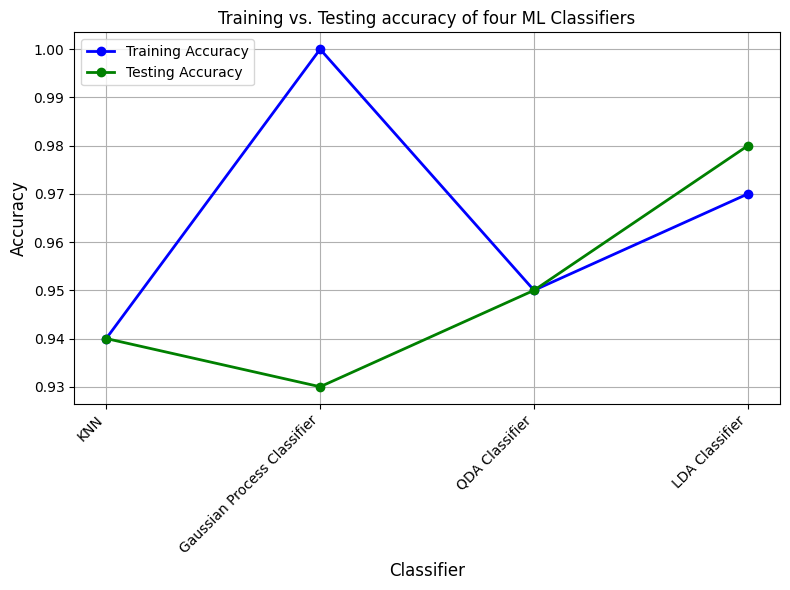}
    \caption{Training vs. Testing Accuracy of Four ML Classifiers.}
    \label{fig:train_test_accuracy}
\end{figure}

Figure \ref{fig:train_test_accuracy} compares the training and testing accuracy of the four classifiers. The Gaussian Process Classifier achieves 100\% training accuracy, which is indicative of overfitting, as its testing accuracy drops to 93\%. In contrast, LDA consistently maintains both high training accuracy (97\%) and high testing accuracy (98\%), showing that it generalizes well to unseen data. This comparison highlights LDA as the most robust model for dementia prediction.

Moreover, Table \ref{tab:evaluation_metrics} presents the precision, recall, and F1-score for each classifier across two classes (class 0 and class 1). LDA delivers outstanding results, with precision and recall values of 1.00 and 0.95 for class 1, resulting in an F1-score of 0.97, the highest among the four models. QDA and KNN follow with slightly lower scores. This table highlights LDA’s ability to accurately predict dementia cases while minimizing both false positives and false negatives.

\begin{table}[h]
\centering
\caption{Evaluation Metrics of Four ML Classifiers}
\label{tab:evaluation_metrics}
\begin{tabular}{p{1.0cm}p{1.0cm}p{1.5cm}p{1.0cm}p{1.0cm}}  % Adjusted column widths
\toprule
\textbf{Model}  & \textbf{Class} & \textbf{Precision} & \textbf{Recall} & \textbf{F1-Score} \\ 
\midrule
KNN    & Class 0 & 0.91  & 0.99  & 0.95 \\ 
       & Class 1 & 0.99  & 0.89  & 0.93 \\ 
\midrule
GPC    & Class 0 & 0.91  & 0.98  & 0.94 \\ 
       & Class 1 & 0.98  & 0.89  & 0.93 \\ 
\midrule
QDA    & Class 0 & 0.94  & 0.99  & 0.96 \\ 
       & Class 1 & 0.99  & 0.93  & 0.96 \\ 
\midrule
LDA    & Class 0 & 0.96  & 1.00  & 0.98 \\ 
       & Class 1 & 1.00  & 0.95  & 0.97 \\ 
\bottomrule
\end{tabular}
\end{table}

\begin{figure}[h!]
    \centering
    \includegraphics[width=0.5\textwidth]{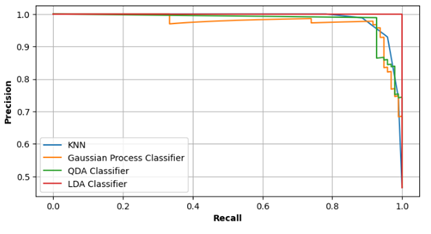}
    \caption{Precision-Recall Curves of Four ML Classifiers.}
    \label{fig:precision_recall}
\end{figure}

Figure \ref{fig:precision_recall} presents the precision-recall curves, which demonstrate the trade-off between precision and recall for the classifiers. LDA exhibits a perfect precision-recall balance with a precision of 1.0 and a recall of 1.0, indicating that it accurately predicts positive dementia cases with minimal false negatives. The other classifiers, while still performing well, do not achieve this ideal balance, further emphasizing the effectiveness of LDA in dementia prediction.

\begin{figure}[h!]
    \centering
    \includegraphics[width=0.4\textwidth]{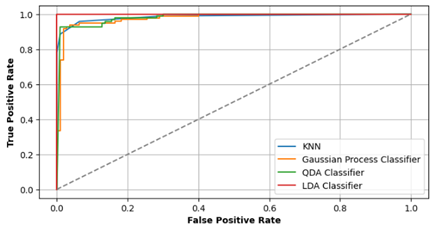}
    \caption{ROC curve of four ML Classifiers.}
    \label{fig:roc_curve}
\end{figure}

The ROC curves in Figure \ref{fig:roc_curve} compare the true positive rate (sensitivity) against the false positive rate for each classifier. The closer the curve is to the top-left corner, the better the model. LDA has the highest area under the curve (AUC), indicating that it is the most optimal classifier for this dataset. The other models perform well but are not as close to the ideal point as LDA.

\begin{figure}[h!]
    \centering
    \includegraphics[width=0.5\textwidth]{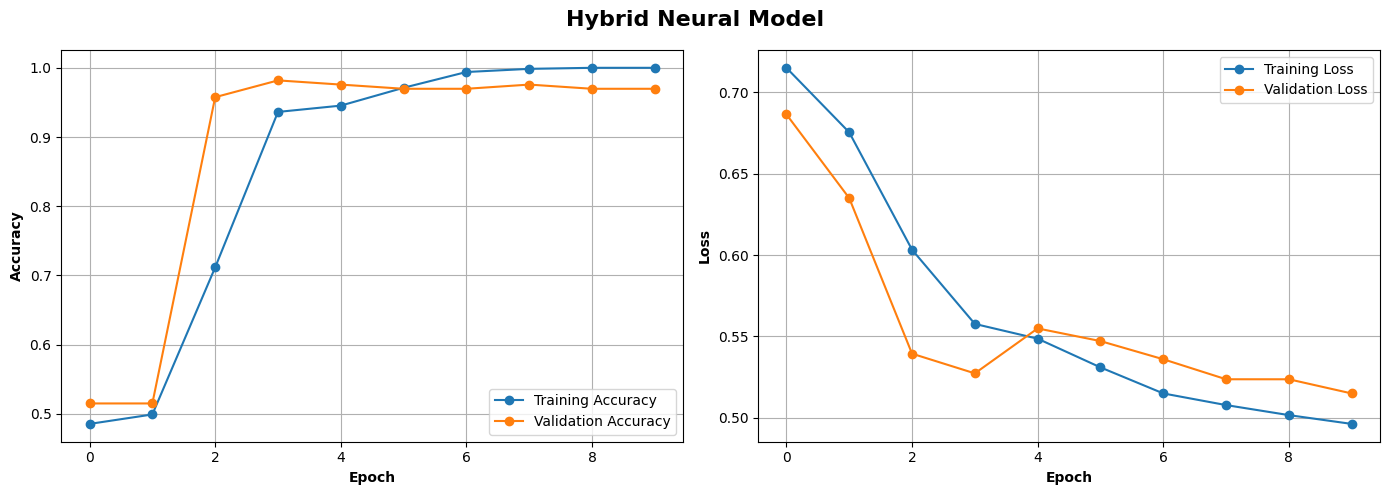}
    \caption{Accuracy and Loss of Hybrid Neural Model.}
    \label{fig:accuracy_loss}
\end{figure}

Figure \ref{fig:accuracy_loss} shows the training and validation accuracy and loss of the hybrid CNN-RNN model over multiple epochs. The training accuracy reaches 100\% by the 8th epoch, while validation accuracy stabilizes at around 97\%. The loss decreases consistently, indicating that the model is learning effectively without overfitting. Despite slightly lower validation accuracy compared to traditional ML models, the hybrid model demonstrates the potential for integrating numerical and text data in dementia prediction.

\begin{figure}[h!]
    \centering
    \includegraphics[width=0.5\textwidth]{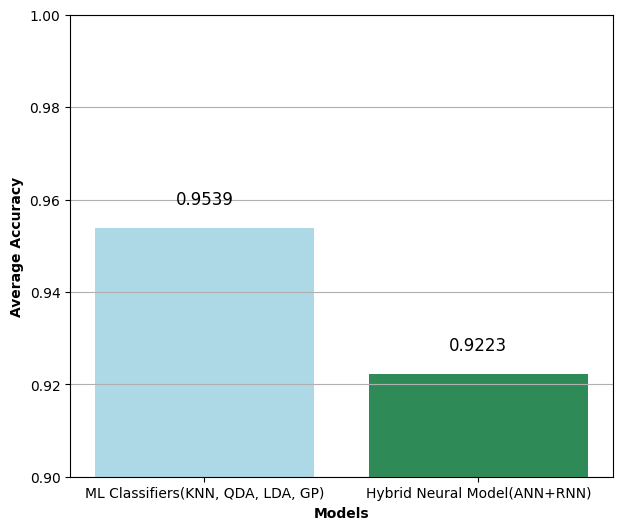}
    \caption{Comparison between ML and Deep Learning Model accuracy.}
    \label{fig:ml_vs_dl}
\end{figure}

Figure \ref{fig:ml_vs_dl}, a comparison is made between the average accuracy of the four ML models and the hybrid deep learning model. The ML models collectively achieve an average accuracy of 95.39\%, outperforming the hybrid CNN-RNN model, which has an accuracy of 92.23\%. This comparison underscores the strength of traditional ML models in this context, though the hybrid model still offers promise for future improvements.

\begin{figure}[h!]
    \centering
    \includegraphics[width=0.5\textwidth]{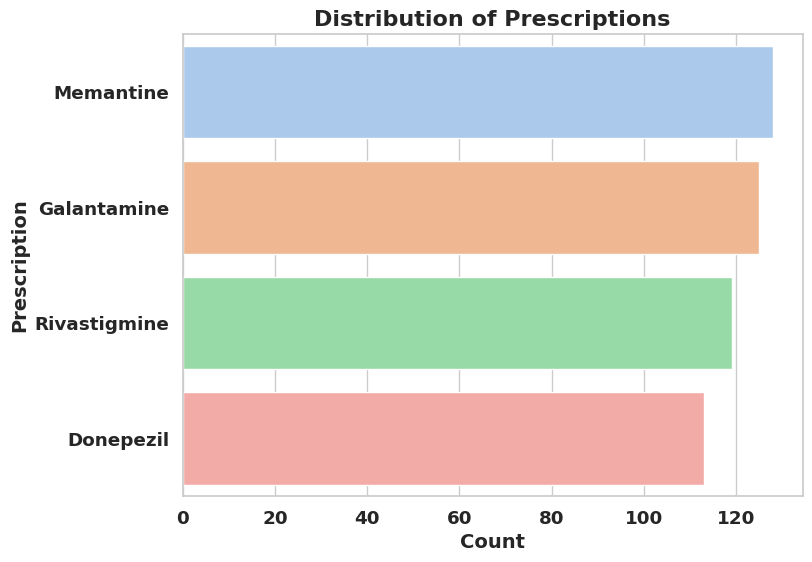}  % Corrected the includegraphics command
    \caption{Prescription Distribution for Dementia patients.}
    \label{fig:prescription}
\end{figure}

Figure \ref{fig:prescription} visualizes the distribution of prescribed medications for dementia patients in the dataset. Memantine is the most commonly prescribed drug, known for its neuroprotective effects in Alzheimer’s disease and other neurodegenerative conditions. Galantamine follows, a drug that enhances cognitive function. Rivastigmine and donepezil, both frequently prescribed, also target cognitive symptoms. The figure suggests that certain medications are more likely to be recommended for patients at higher risk of dementia based on ML models’ evaluations of patient health data.

\section{DISCUSSION}

The discussion of this manuscript emphasizes the significant advancements made in dementia prediction through the application of various ML techniques. The study highlights the efficacy of supervised learning models, such as KNN, QDA, LDA, and Gaussian Process Classifiers, in accurately predicting dementia. Notably, the research demonstrates that LDA outperforms other classifiers with an impressive testing accuracy of 98\%. This underscores the potential of ML in enhancing early detection and management of dementia, thus enabling timely interventions. 

The research also explores the integration of advanced feature engineering techniques like TF-IDF vectorization and the use of SMOTE to address class imbalance, further boosting the performance of ML models. The hybrid neural network model, combining CNN and Recurrent Neural Networks (RNN), although slightly less accurate than traditional ML models, presents a novel approach by leveraging both numerical and textual data for dementia prediction.

Additionally, the study underscores the importance of interpretability in ML models, providing insights into how different features contribute to dementia prediction, which is crucial for clinical applications. The findings indicate a strong correlation between certain features, such as APOE-$\varepsilon$4 allele presence and chronic conditions like diabetes, and the likelihood of developing dementia. 

This comprehensive analysis of various ML techniques and their applications in dementia prediction provides valuable insights and practical implications for healthcare providers, reinforcing the potential of ML in transforming dementia care and enhancing the accuracy and reliability of early diagnosis. The study calls for continued innovation and refinement in ML methodologies to further improve the predictive capabilities and clinical utility of these technologies in dementia care.

\section{Future Research Direction}
The future direction of this study will focus on several key advancements to enhance the predictive capabilities and clinical applicability of ML models in dementia detection. Expanding the dataset by incorporating more diverse and comprehensive patient data from various sources, including genetic, imaging, and clinical data, is crucial for improving model generalizability and robustness. Additionally, exploring advanced ML techniques such as deep learning and ensemble methods can further enhance prediction accuracy and reliability. Integrating explainable AI approaches will be essential to ensure clinicians can understand and trust the models' decision-making processes, facilitating their adoption in clinical settings. Longitudinal studies should be conducted to evaluate the models' performance over time and their ability to predict the progression of dementia, providing valuable insights for early intervention and personalized treatment plans. Finally, collaboration with healthcare providers and researchers will be pivotal in refining these models and implementing them in real-world clinical practices to improve dementia care and outcomes.

\section{CONCLUSION}

This study underscores the transformative potential of ML in the early detection and management of dementia, with LDA achieving the highest testing accuracy of 98\%. By employing advanced feature engineering techniques such as TF-IDF vectorization and SMOTE, the study significantly enhanced model performance. Additionally, the introduction of a hybrid neural network model combining CNN and Recurrent Neural Networks (RNN) demonstrated a novel approach, though slightly less accurate than traditional ML models. Key findings highlight strong correlations between specific features like the APOE-$\epsilon$4 allele and chronic conditions such as diabetes, emphasizing the importance of genetic and health data integration in predictive models. The study also stresses the critical role of model interpretability in clinical applications, providing valuable insights for healthcare providers. In conclusion, our research advocates for continued innovation in ML methodologies to further improve their predictive capabilities and clinical utility, revolutionizing dementia diagnosis and management through early detection and timely interventions.


\begin{thebibliography}{99}

\bibitem{ref1} Kumar, S., Oh, I., Schindler, S., Lai, A. M., Payne, P. R., \& Gupta, A. (2021). Machine learning for modeling the progression of Alzheimer disease dementia using clinical data: a systematic literature review. \textit{JAMIA open}, 4(3), ooab052.

\bibitem{ref2} Harshfield, E. L., Shi, L., Badhwar, A., Khleifat, A. A., Clarke, N., Dehsarvi, A., \ldots \& Winchester, L. M. (2023). Artificial intelligence for biomarker discovery in Alzheimer's disease and dementia.

\bibitem{ref3} Rajayyan, S., \& Mustafa, S. M. M. (2023). Prediction of dementia using machine learning model and performance improvement with cuckoo algorithm. \textit{International Journal of Electrical \& Computer Engineering}, 13(4).

\bibitem{ref4} Park, C., Jang, J. W., Joo, G., Kim, Y., Kim, S., Byeon, G., \ldots \& Kim, S. (2022). Predicting progression to dementia with “comprehensive visual rating scale” and machine learning algorithms. \textit{Frontiers in Neurology}, 13, 906257.

\bibitem{ref5} Newby, D., Orgeta, V., Marshall, C. R., Lourida, I., Albertyn, C. P., Tamburin, S., \ldots \& Ranson, J. M. (2023). Artificial intelligence for dementia prevention. \textit{Alzheimer's \& Dementia}, 19(12), 5952-5969.

\bibitem{ref6} Nazir, A. (2019). A critique of imbalanced data learning approaches for big data analytics. \textit{International Journal of Business Intelligence and Data Mining}, 14(4), 419-457.

\bibitem{ref7} Khan, A., Zubair, S., \& Khan, S. (2022). A systematic analysis of assorted machine learning classifiers to assess their potential in accurate prediction of dementia. \textit{Arab Gulf Journal of Scientific Research}, 40(1), 2-24.

\bibitem{ref8} Li, W., Zeng, L., Yuan, S., Shang, Y., Zhuang, W., Chen, Z., \& Lyu, J. (2023). Machine learning for the prediction of cognitive impairment in older adults. \textit{Frontiers in Neuroscience}, 17, 1158141.

\bibitem{ref9} Chaki, J., \& Woźniak, M. (2023). Deep learning for neurodegenerative disorder (2016 to 2022): A systematic review. \textit{Biomedical Signal Processing and Control}, 80, 104223.

\bibitem{ref10} Herzog, N. J. (2023). Deep learning of brain asymmetry digital biomarkers to support early diagnosis of cognitive decline and dementia (Doctoral dissertation, Birkbeck, University of London).

\bibitem{ref11} Herzog, N. J., \& Magoulas, G. D. (2022). Convolutional neural networks-based framework for early identification of dementia using MRI of brain asymmetry. \textit{International Journal of Neural Systems}, 32(12), 2250053.

\bibitem{ref12} Irfan, M., Shahrestani, S., \& Elkhodr, M. (2023). Enhancing Early Dementia Detection: A Machine Learning Approach Leveraging Cognitive and Neuroimaging Features for Optimal Predictive Performance. \textit{Applied Sciences}, 13(18), 10470.

\bibitem{ref13} Rajayyan, S., \& Mustafa, S. M. M. (2023). Prediction of dementia using machine learning model and performance improvement with cuckoo algorithm. \textit{International Journal of Electrical \& Computer Engineering}, 13(4).

\bibitem{ref14} Vyshnavi, P., Challagulla, S. P., Adamu, M., Vicencio, F., Jameel, M., Ibrahim, Y. E., \& Ahmed, O. S. (2023). Utilizing Artificial Neural Networks and Random Forests to Forecast the Dynamic Amplification Factors of Non-Structural Components. \textit{Applied Sciences}, 13(20), 11329.

\bibitem{ref15} Kumkum, A. R., Sen, A., Noman, A. Y., Majumder, P., Liew, T. H., Fahad, N., Miah, M. S. U., Rabbi, R. I., \& Hossen, M. J. (2025). From data to diagnosis: Applying machine learning model for reliable heart disease prediction. 2025 5th Asia Conference on Information Engineering (ACIE), 50-56. https://doi.org/10.1109/ACIE64499.2025.00015

\bibitem{ref16} Dhakal, S., Azam, S., Hasib, K. M., Karim, A., Jonkman, M., \& Al Haque, A. F. (2023). Dementia prediction using machine learning. \textit{Procedia Computer Science}, 219, 1297-1308.

\bibitem{ref17} Bidani, A., Gouider, M. S., \& Travieso-González, C. M. (2019). Dementia detection and classification from MRI images using deep neural networks and transfer learning. In \textit{Advances in Computational Intelligence: 15th International Work-Conference on Artificial Neural Networks, IWANN 2019, Gran Canaria, Spain, June 12-14, 2019, Proceedings, Part I 15} (pp. 925-933). Springer International Publishing.

\bibitem{ref18} Fahad, N., Sen, A., Jisha, S. S., Ahmad, S., Mokhlis, H., \& Hossain, M. S. (2023, December). Identification of human movement through a novel machine learning approach [Paper presentation]. 2023 Innovations in Power and Advanced Computing Technologies (i-PACT), Kuala Lumpur, Malaysia. https://doi.org/10.1109/i-PACT58649.2023.10434296

\bibitem{ref19} Ahmed, R. (2025). \textit{dementia\_prediction} [Dataset]. GitHub. \url{https://github.com/raselahmed1337/dementia_prediction}. Accessed on December 5, 2024.

\bibitem{ref20} Rawat, R. M., Akram, M., \& Pradeep, S. S. (2020, June). Dementia detection using machine learning by stacking models. In \textit{2020 5th International Conference on Communication and Electronics Systems (ICCES)} (pp. 849-854). IEEE.

\bibitem{ref21} Javeed, A., Dallora, A. L., Sanmartin Berglund, J., Ali, A., Ali, L., \& Anderberg, P. (2023). Machine learning for dementia prediction: a systematic review and future research directions. \textit{Journal of Medical Systems}, 47(1), 17.

\end{thebibliography}
\end{document}